\documentclass[twocolumn]{NobArticle}
\runninghead{Mercury by Hypernym AI}
\footertext{\textit{Technical White Paper} (May 2025)}

\title{{\color[RGB]{164,27,27}{H}}%
       {\color[RGB]{247,185,121}{Y}}%
       {\color[RGB]{196,153,21}{P}}%
       {\color[RGB]{68,126,42}{E}}%
       {\color[RGB]{85,140,152}{R}}%
       {\color[RGB]{81,135,220}{N}}%
       {\color[RGB]{167,202,234}{Y}}%
       {\color[RGB]{59,46,98}{M}}
       {\color{magenta}{M}}{\color{orange}{E}}{\color{pink}{R}}{\color{teal}{C}}{\color{cyan}{U}}{\color{purple}{R}}{\color{violet}{Y}}: Token Opti{\color{magenta}{m}}ization through S{\color{orange}{e}}mantic Field Const{\color{pink}{r}}iction and Re{\color{teal}{c}}onstr{\color{cyan}{u}}ction from Hype{\color{purple}{r}}n{\color{violet}{y}}ms. A New Text Compression Method}

\newcommand{\subtitle}[1]{Mercury by Hypernym AI}

\author{
     Chris Forrester\textsuperscript{$\ast $}  and
     Octavia Sulea\textsuperscript{$\infty $}
}

\date{
    \textsuperscript{{$\ast $}}
    chris@hypernym.ai \\
    \textsuperscript{{$\infty $}}
    octavia@hypernym.ai
}

\renewcommand{\maketitlehookd}{%
\begin{abstract}
    Compute optimization using token reduction of LLM prompts is an emerging task in the fields of NLP and next generation, agentic AI. In this white paper, we introduce a novel (patent pending) text representation scheme and a first-of-its-kind word-level semantic compression of paragraphs that can lead to over 90\% token reduction, while retaining high semantic similarity to the source text. We explain how this novel compression technique can be lossless and how the detail granularity is controllable. We discuss benchmark results over open source data (i.e. Bram Stoker's \textit{Dracula} available through Project Gutenberg) and show how our results hold at the paragraph level, across multiple genres and models. 
    \small{\textbf{Index Terms:} Token Optimization, Text Representation, Semantic Compression}
\end{abstract}
}

\begin{document}

\small
\maketitle


\section{Introduction}
Token consumption and computation costs are predicted to increase exponentially as LLM-based agent applications continue to take over the AI market. Adoption of \textit{agentic AI} is predicted to increase drastically by the end of this year, with hyper-personalization being the next target on the way to SGI. While hardware strategies are being investigated to mitigate and capture carbon emissions from GPU usage, a new task in classical NLP is emerging\footnote{\url{https://tinyurl.com/microsoft-tech-blog}} \footnote{\url{https://tinyurl.com/ibm-developer-blog}}: token optimization. In this white paper, we present a new (patent-pending) text representation technique that draws concepts from the fields of data compression, information theory, game theory, and NLP to arrive at a first-of-its-kind word-level semantic text compression algorithm. This algorithm is available for public and commercial use through the Hypernym API, under the product name Mercury\footnote{\url{https://www.hypernym.ai/pricing}}.   

\subsection{Contribution}
This white paper has the following contributions.

\begin{itemize}
	\item Introduces of a novel (patent pending) semantic representation of text and a first-of-its-kind word-level \textit{text compression} technique;
	\item Shows how this technique leads to significant decrease in and controllable token utilization of LLM API calls; 
	\item Shows how this compression can be lossless and that it behaves consistently across the LLM spectrum in RAG applications;
	\item Establishes new baselines for the state of the art in text encoding
\end{itemize}


\section{Background}

Recently, knowledge distillation has been proposed in \cite{hu2023kd} as a technique by which a smaller model (the student) learns to mimic the behavior of a larger model (the teacher). The smaller model is trained on the soft labels (logits) of the larger model. This ties the problem of token optimization to that of optimizing the size of the model parameters. Around the same time, in computer vision and image generation, token optimization started to be approached by merging similar tokens \cite{bolya2023}. In \cite{gilbert2023}, the capabilities of LLMs to perform semantic compression over the prompt text are investigated. This paper falls into the larger NLP subfield of modern abstractive summarization, where systems aim to retain overall meaning and the most salient details, often using techniques like generalization and hypernyms to replace specific details with more general terms. These systems, while powerful, can be unpredictable. They typically lack a guarantee of semantic fidelity. They may drop details that seem minor but are contextually important, and they do not provide a mechanism to recover or explicitly include those details in a structured way. No known summarization model uses a game-theoretic importance metric or a dedicated representational structure like the one we introduce here: \textit{the dart}. Furthermore, pure neural summaries do not include a verification step using independent models, which means errors or hallucinations can slip through unnoticed.
\section{Methods}
At a high level, the Hypernym compression process involves the following stages:
\begin{enumerate}

    \item Parsing and Dart Structuring: The input text is analyzed linguistically and semantically to identify its general concepts and specific details. The content is then represented in a structured format dubbed \textit{a dart}, which serves as an intermediate representation for compression;
   \item Detail Importance Evaluation (Game-Theoretic Analysis): Each detail in the dart is evaluated for its contribution to the overall meaning using Shapley value calculations derived from cooperative game theory. This yields an importance score for every detail.
  \item Detail Swapping and Compression Optimization: Guided by importance scores, the system iteratively removes or replaces less critical details to achieve compression while attempting to keep the meaning unchanged. This may involve swapping a detail with a more general descriptor (hypernym) plus a minimal indicator of the original detail, or other transformations.
  \item Multi-model Verification: The compressed output is then fed through multiple independent models or analysis modules to verify that it maintains semantic fidelity. If any model detects a significant deviation in meaning or key information, the system can adjust (e.g., reinstate a dropped detail or try an alternative compression for that segment).
  \item	Output generation (Dart to Text): Finally, the verified dart structure is converted back into a human-readable compressed text format. The end result is a summary that is highly concise, but preserves the essential field of information from the original – hence the name Field Constrictor - as it restricts the field of data without strangling its meaning.
\end{enumerate}

\subsection{Dart Structure}
The dart is a novel semantic representation for text. In a dart structure, a piece of information is split in two parts: 
\begin{itemize}
    \item a \textbf{core} statement that uses generalized terms (such as hypernyms or abstracted descriptions) to represent the main idea and 
    \item a set of \textbf{details} that provide specific information (that would normally make the text longer if included inline).
\end{itemize}

The structure is analogous to a dart with a tip and a tail: the \textit{tip} is the concise core (pointing to the main meaning), and the \textit{tail} carries the details in a compact form. For example, consider an original sentence: “The German shepherd named Rex barked loudly at the mail carrier at 7:00 AM.” A dart-structured representation could be: “A dog barked loudly at a mail carrier. (Details: breed=German Shepherd, name=Rex, time=7:00 AM).”** Here, the core statement uses a hypernym “dog” (a general category) instead of the specific “German shepherd named Rex”, and omits the exact time. The details that were removed from the main sentence are appended in a structured way (“breed=… , name=…, time=…”). This retains the key specifics in a very compressed metadata-like form.

\subsection{Field Constrictor}
The Field Constrictor system automatically generates such dart structures. The algorithm identifies when a specific entity or detail can be replaced by a broader term plus an annotation in the details. The conversion to a dart may involve the following:

\begin{itemize}
    \item Replacing named entities or proper nouns with their category (i.e. hypernym). (E.g., person’s name -> “a person”; a specific model of car -> “a car”.);
    \item Abstracting numerical or granular information into higher-level terms, with specifics moved to details. (E.g., converting “7:00 AM” to “early morning” in the dart core statement, but noting the exact time in the details.);
    \item Maintaining relational or action information in the core so the sentence remains coherent and grammatically correct.
\end{itemize}

\subsection{Recomposition}
We use the final (hypernym-abstracted) representation and several of the popular LLMs to reconstruct (generate) the original paragraph at desired granularity. The generated darts are also saved for different-fidelity reconstructions. We compare the reconstructed with the field-constricted representations for semantic similarity under different compression rates. 


\section{Experimental Results}

We performed benchmark experiments using the classic Gutenberg book, \textit{Dracula}, and a straight-forward (TF-IDF retrieval) RAG engine. Additionally, we performed case studies at the paragraph level, to analyze different passages pertaining to different genres and styles. The paragraphs for each type of passage (representing the top five most difficult genres to summarize) are given in full in the Appendix.

In both sets of studies, we followed the steps below.

\begin{itemize}
\item Prepare the Original paragraph into chunks through the Unstructured API \footnote{\url{https://docs.unstructured.io/open-source}};
\item Create Hypernym API calls with these chunks - this gives a compatibility metric;
\item On RAG retriever match, recall the Original chunks or Hypernym chunks based on desired compatibility threshold.
\end{itemize}

\begin{figure}[!htpb]
    \centering
    \includegraphics[width=\linewidth]{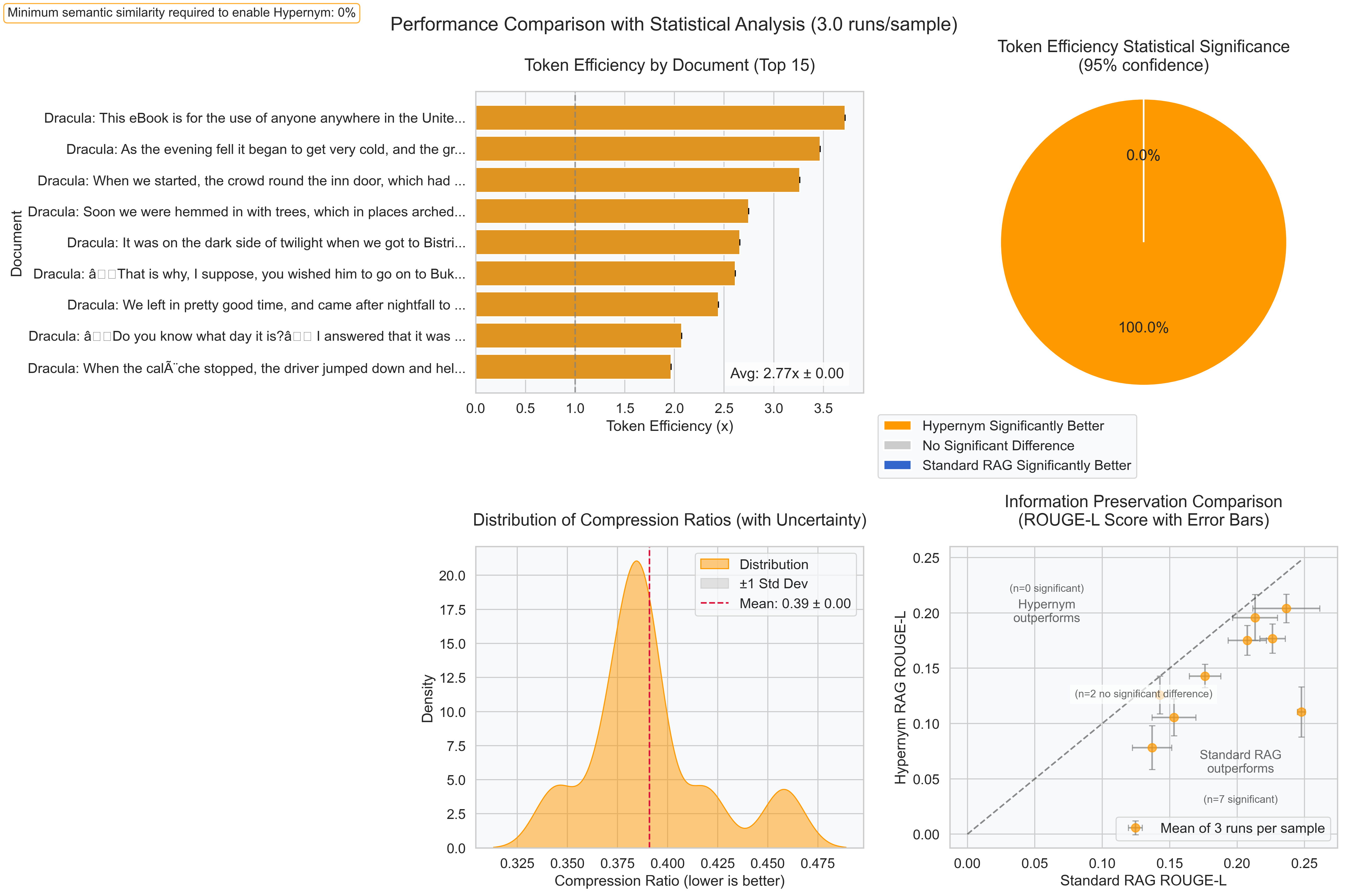}
    \caption{Token optimization for RAG over Dracula}
    \label{fig:dracula}
\end{figure}

In figure \ref{fig:dracula}, we illustrate the token efficiency of the top 15 paragraphs (upper left corner), the statistical significance of the token efficiency measurements throughout the book (upper right corner), the distribution of compression ratios (lower left corner) and the comparison of the ROUGE-L score between standard and Hypernym-RAG (lower right corner). For a more detailed analysis of these results, see our dedicated benchmark report page\footnote{\url{http://max.hypernym.ai/}}.

In the following figures, we note a trend regardless of model chosen for constriction and reconstruction: the third and fourth passages are consistently harder to compress while retaining the same level of semantic similarity with the original. Note that, here, "passage" and "paragraph" are used interchangeably, for brevity: we applied the dart construction part of the field constriction algorithm 60 times for each paragraph of each passage (see the Appendix). For a custom paragraph test, visit our MCP compressor page\footnote{\url{https://tinyurl.com/hypernym-mercury}}.

\begin{figure}[!htpb]
    \centering
    \includegraphics[width=\linewidth]{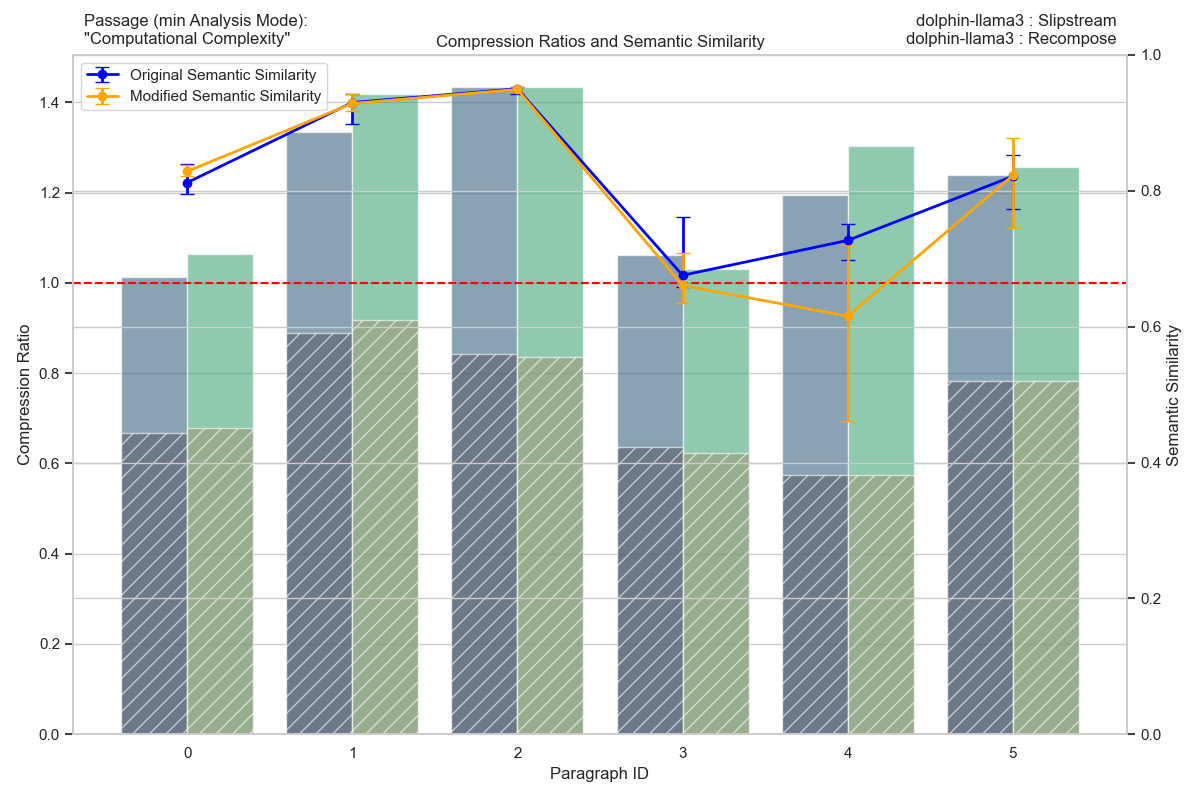}
    \caption{Field constriction (slipstream) and reconstruction using dolphin-llama3 }
    \label{fig:llama3-llama3}
\end{figure}

\begin{figure}[!htpb]
    \centering
    \includegraphics[width=\linewidth]{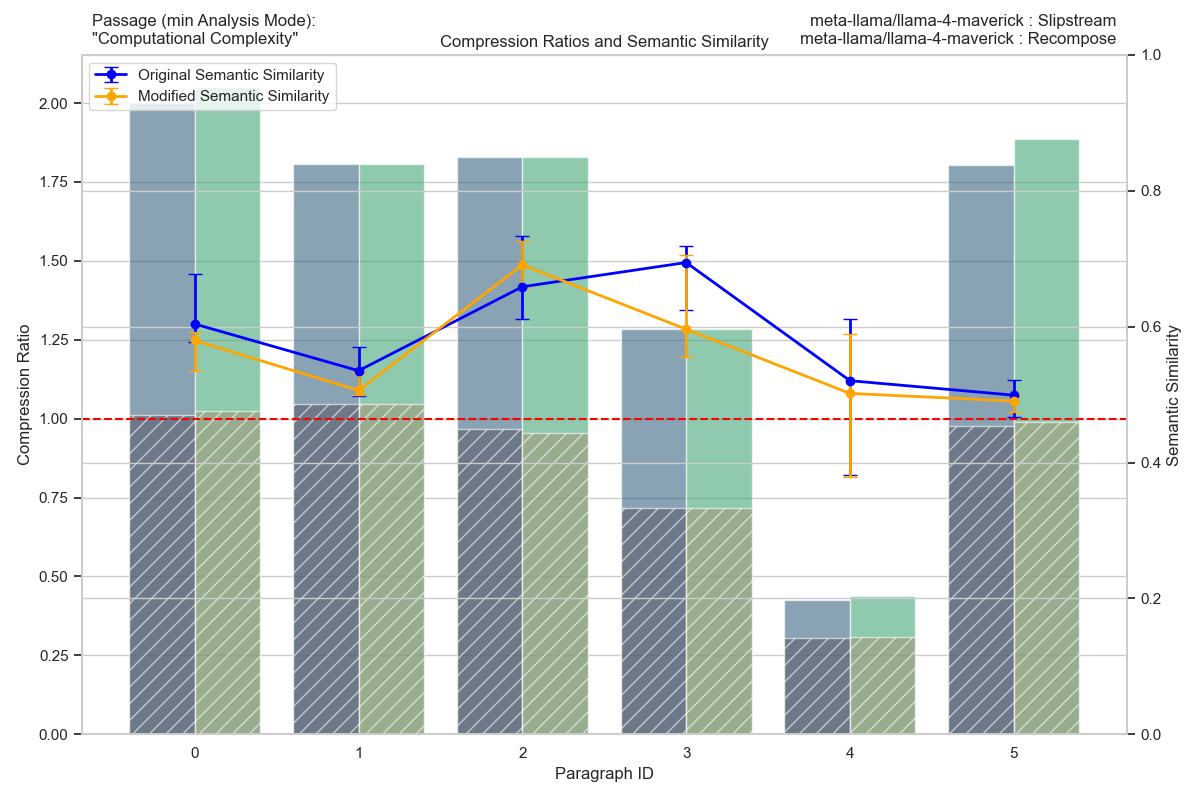}
    \caption{Field constriction and reconstruction using llama4-maverick}
    \label{fig:llama4-llama4}
\end{figure}

\begin{figure}[!htpb]
    \centering
    \includegraphics[width=\linewidth]{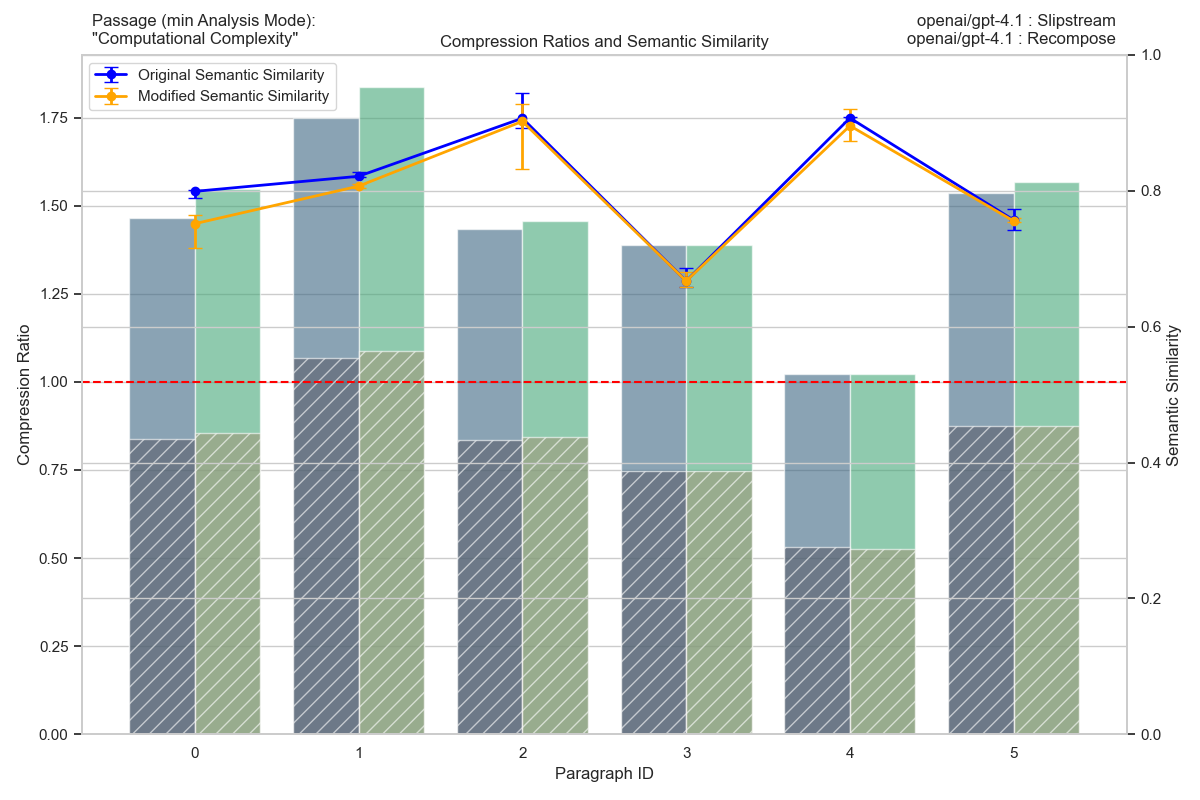}
    \caption{Field constriction and reconstruction using OpenAI's gpt4.1}
    \label{fig:gpt4_1-gpt4_1}
\end{figure}

\begin{figure}[!htpb]
    \centering
    \includegraphics[width=\linewidth]{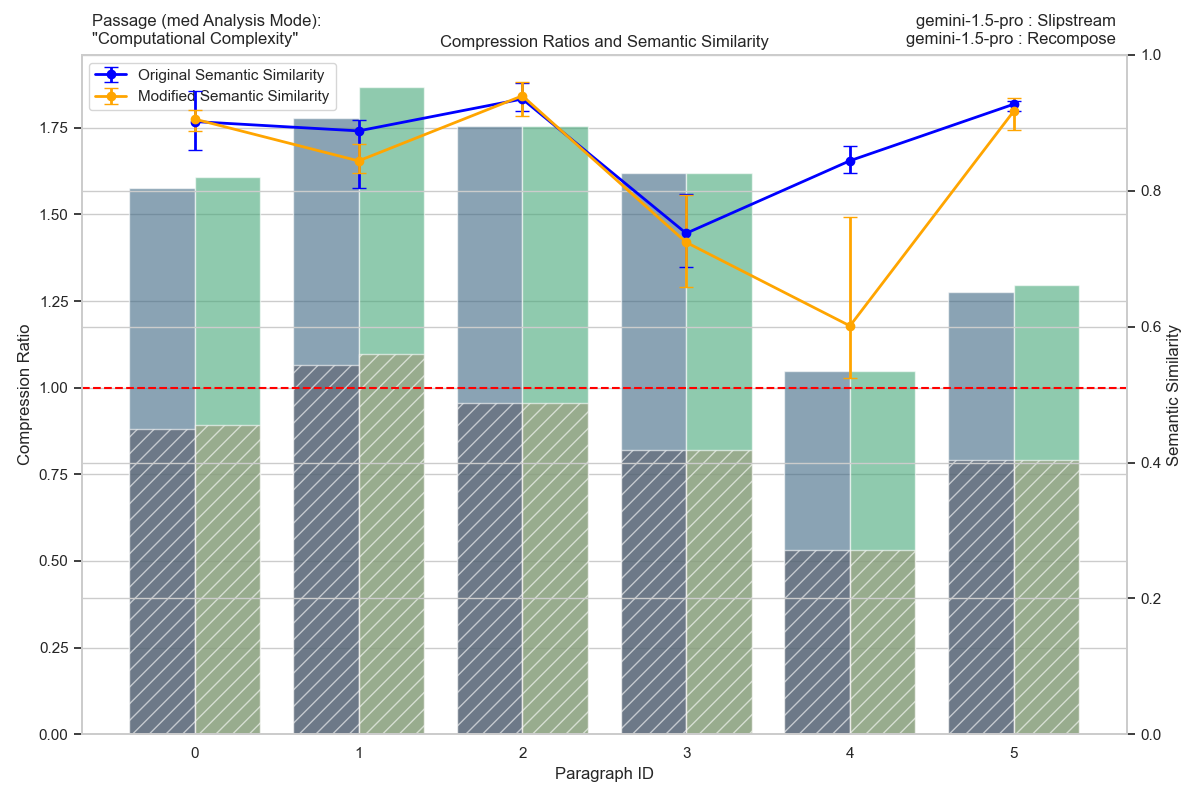}
    \caption{Field constriction and reconstruction using Google's gemini1.5-pro}
    \label{fig:gemini1_5_pro-gemini1_5_pro}
\end{figure}

\begin{figure}[!htpb]
    \centering
    \includegraphics[width=\linewidth]{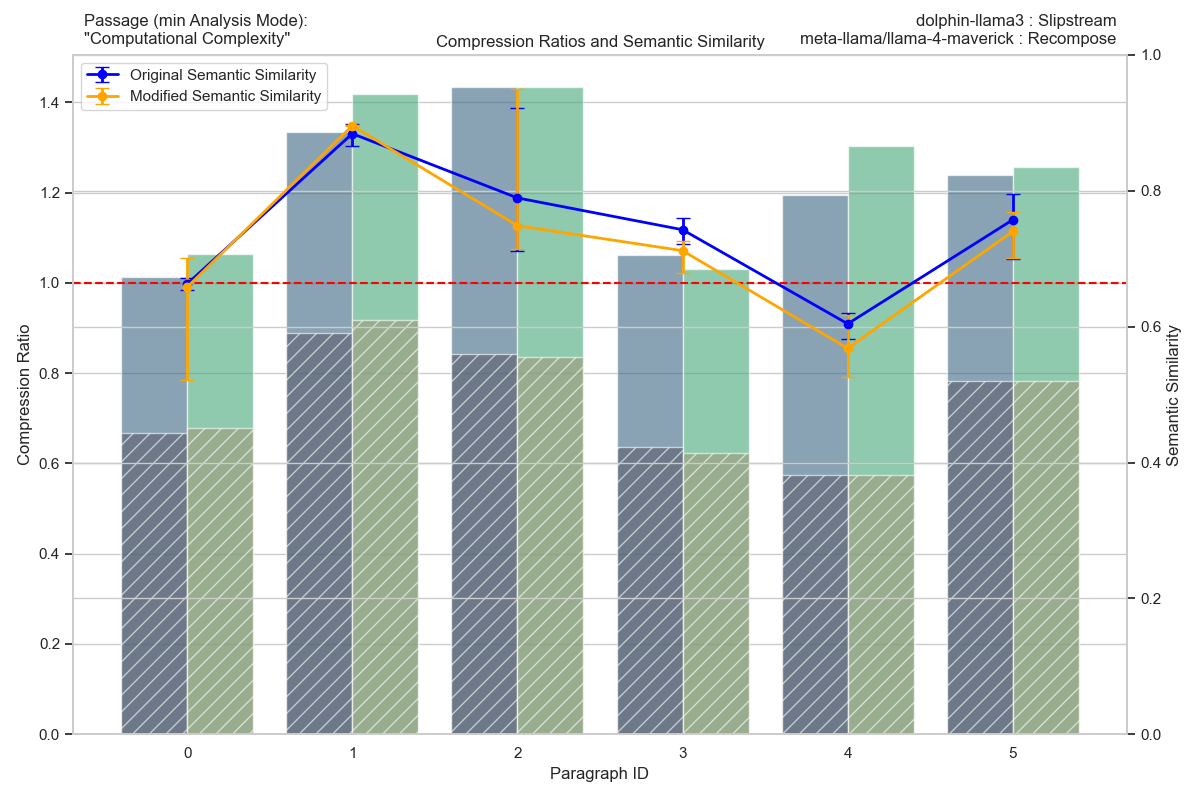}
    \caption{Field constriction using Meta AI's dolphin-llama3 and reconstruction using Meta AI's llama4-maverick}
    \label{fig:llama3-llama4}
\end{figure}

\begin{figure}[!htpb]
    \centering
    \includegraphics[width=\linewidth]{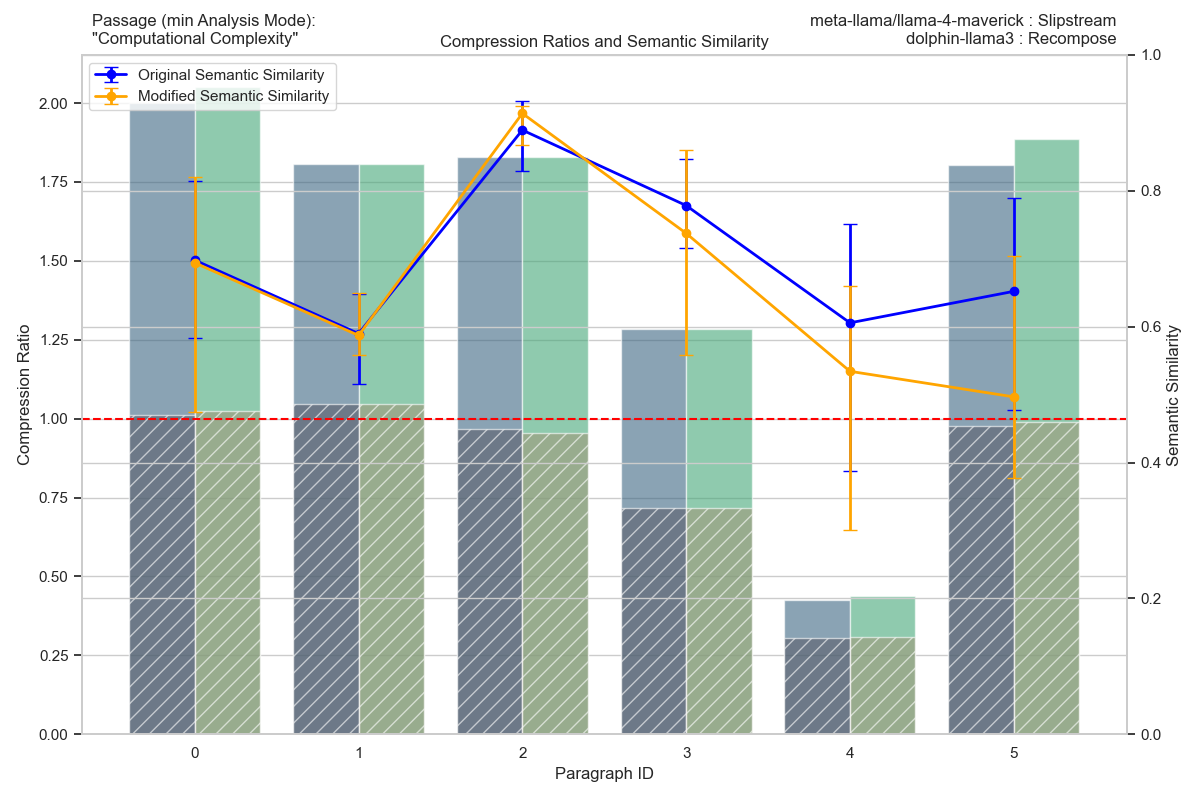}
    \caption{Field constriction using Meta AI's llama4-maverick and reconstruction using Meta AI's dolphin-llama3}
    \label{fig:llama4-llama3}
\end{figure}

\begin{figure}[!htpb]
    \centering
    \includegraphics[width=\linewidth]{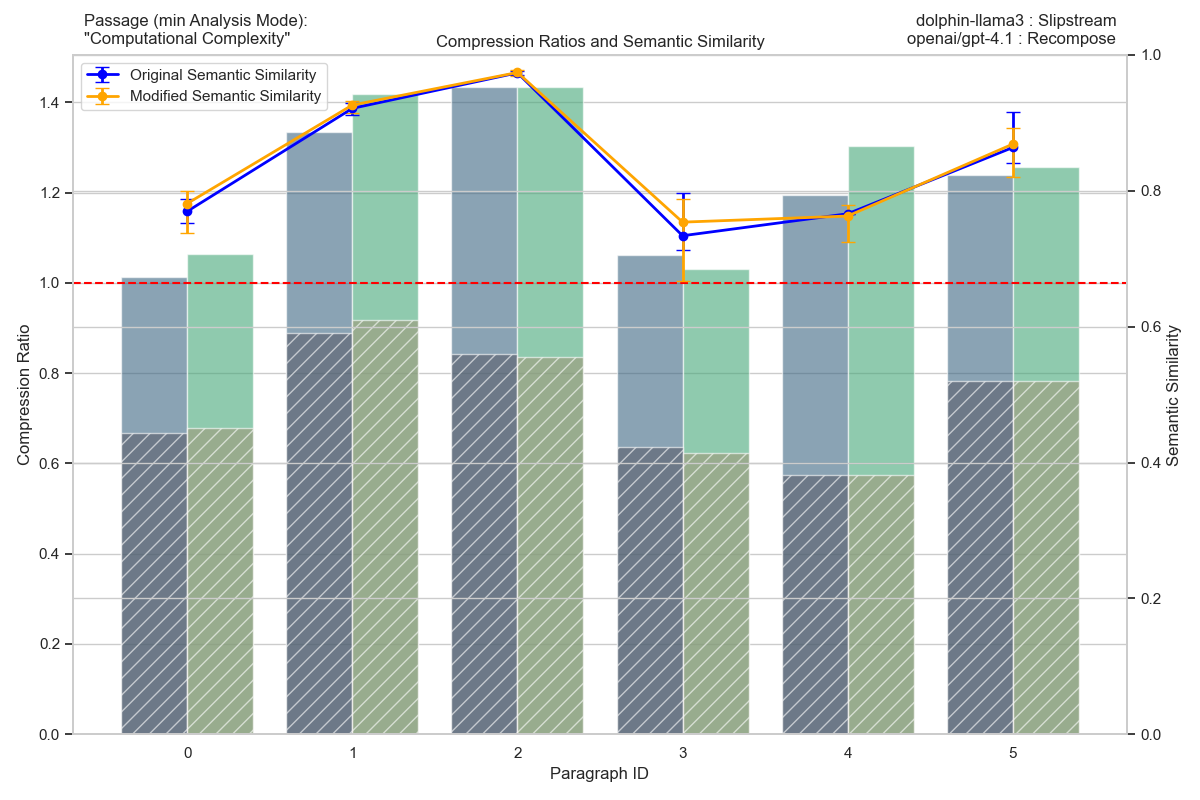}
    \caption{Field constriction using Meta AI's dolphin-llama3 and reconstruction using OpenAI's gpt4.1}
    \label{fig:llama3-gpt4_1}
\end{figure}

\begin{figure}[!htpb]
    \centering
    \includegraphics[width=\linewidth]{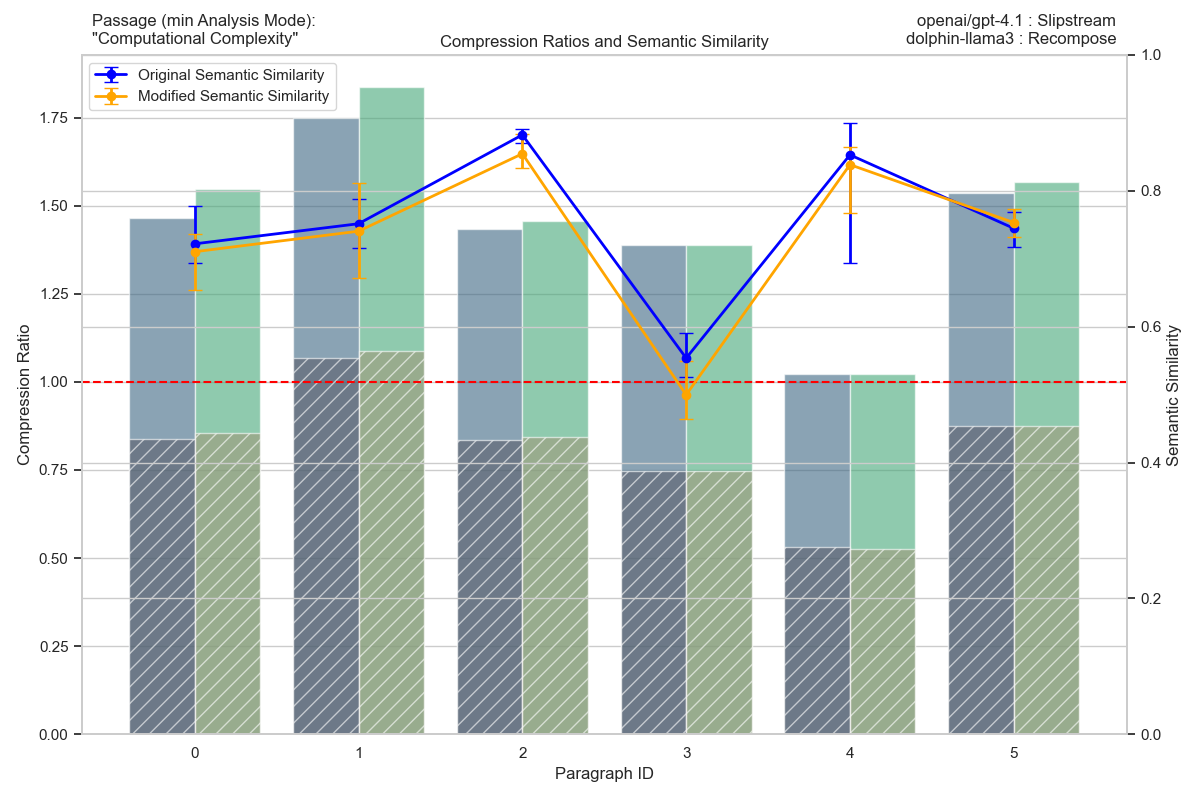}
    \caption{Field constriction using OpenAI's gpt4.1 and reconstruction using Meta AI's dolphin-llama3}
    \label{fig:gpt4_1-llama3}
\end{figure}

\begin{figure}[!htpb]
    \centering
    \includegraphics[width=\linewidth]{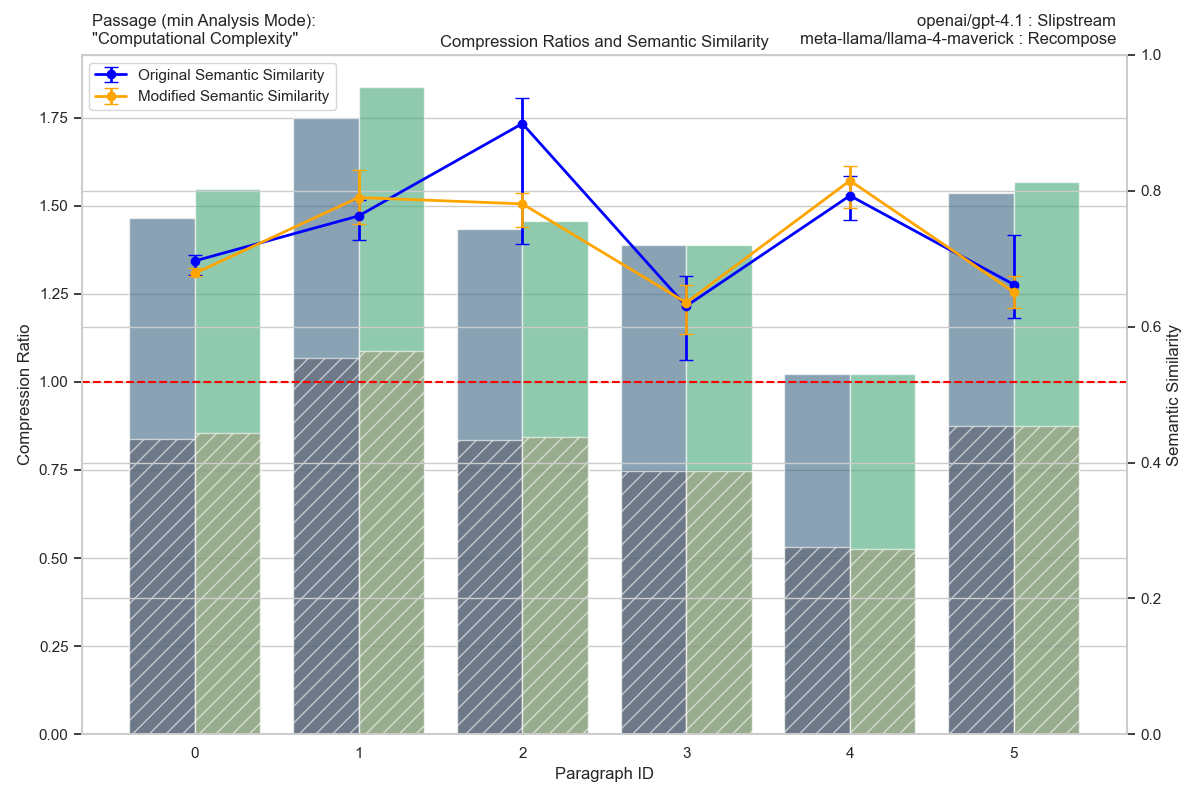}
    \caption{Field constriction using OpenAI's gpt4.1 and reconstruction using Meta AI's llama4-maverick}
    \label{fig:gpt4_1-llama4}
\end{figure}

\begin{figure}[!htpb]
    \centering
    \includegraphics[width=\linewidth]{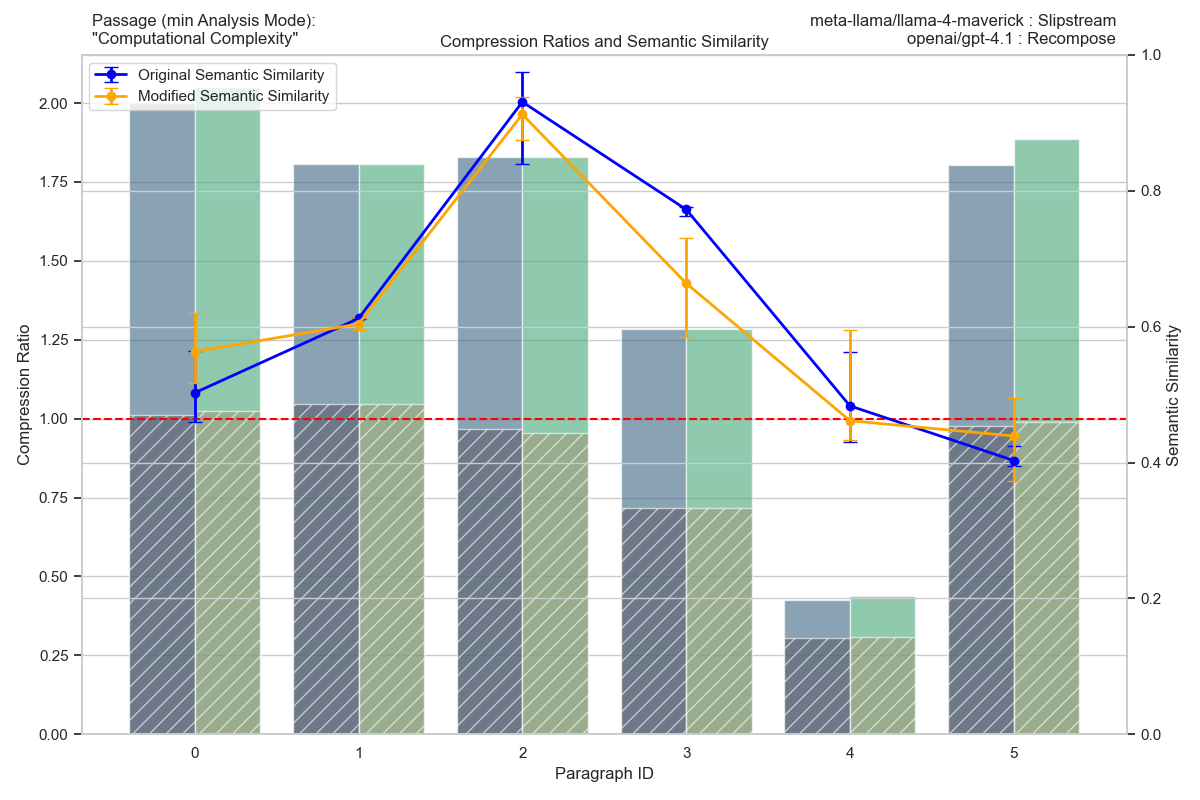}
    \caption{Field constriction using Meta AI's llama4-maverick and reconstruction using OpenAI's gpt4.1}
    \label{fig:llama4-gpt4_1}
\end{figure}

\begin{figure}[!htpb]
    \centering
    \includegraphics[width=\linewidth]{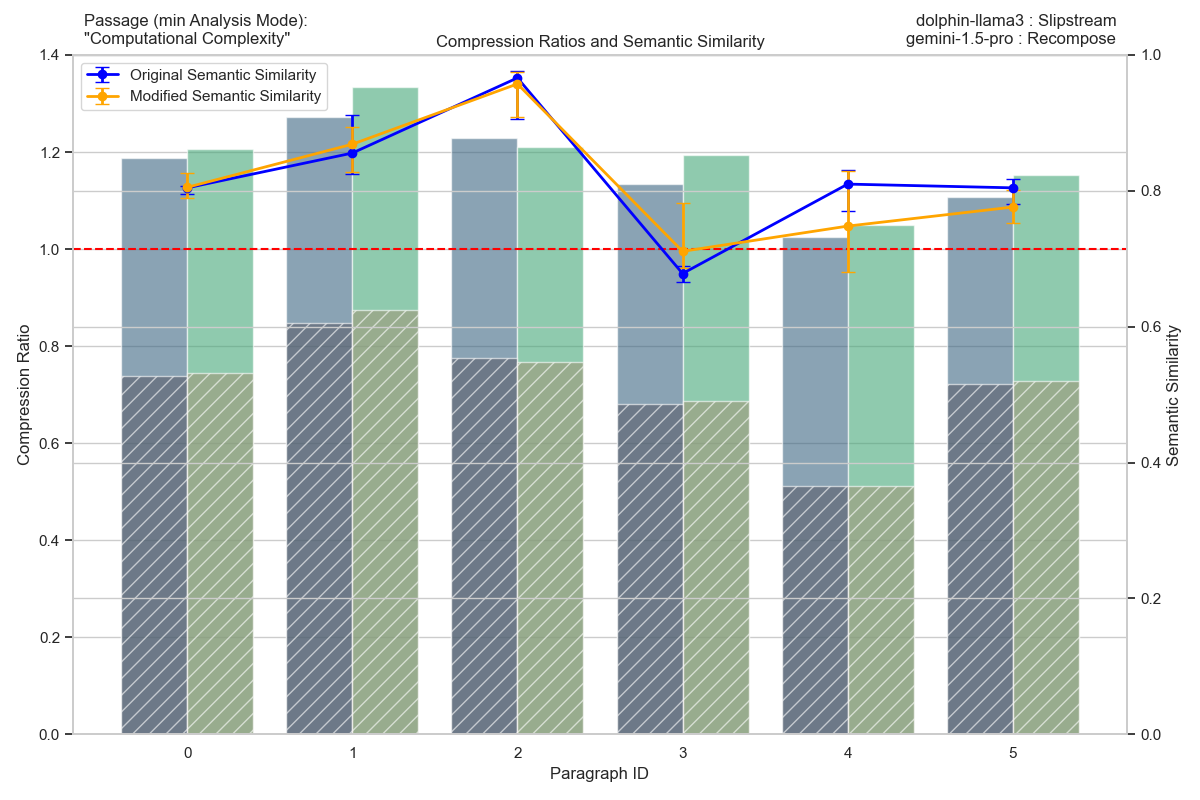}
    \caption{Field constriction using Meta AI's dolphin-llama3 and reconstruction using Google's gemini1.5-pro}
    \label{fig:llama3-gemini1_5_pro}
\end{figure}

\section{Discussion}

Within the logs of our constriction-reconstruction trials, we saw that, while we do reconstruct the wordy-version of a detail which we replace with a simple index during constriction, that wordy-version can act like an intermediate predictive compression verification. For example, in a passage from Dracula relating to scheduling a meeting, we see the following trace emerge in our logs:

"scheduling early morning meeting downtown" = hypernym
[original element]
"time choice"
[replaced element]
"0"
[detail]
"7:00 or 10:00am depending on the weather"

We see this type of trace related to time generalization and reconstruction occurring persistently in the rolling logs. This log trace exemplifies the human readability of the field constriction and reconstruction processes, which are baked in the dart structure. In this case, the reduction of tokens from the specific time choices (e.g. \textit{7:00 or 10:00am depending on the weather}) to a generic (e.g. \textit{0}) saves 20\%-30\% of the total prompt token length in a typical RAG application. This illustrates how our basic premise of predictive encoding becomes fully tangible within the Hypernym AI tool space.

Another observation that emerges from the multiple constriction and reconstruction trials over the various types of passages and models is that, even when the semantic similarity ratio is not good on the first pass, we can find arbitrarily higher numbers within whatever potential well we have established with the hypernym anchor. This isn't just us cherry picking extraordinarily high results that are glitches in any embedding (e.g. BERT, snowflake, etc.), but in fact just potential picks from a nearly endless stream of near-matches that are also internally consistent across all LLM embeddings.

\section{Conclusion}

In this white paper, we have introduced a novel word-level text compression technique that offers a dramatic reduction of tokens in an LLM prompt while maintaining all its saliency. Our results show something beyond the mere surpassing of benchmark SOTA in a given NLP task. Through our novel text compression technology, Mercury, we are defining new horizons in RAG and agentic reasoning that future-proof any text application.

\section*{Acknowledgments}
The authors would like to thank their Hypernym colleagues -Nick, Andy, Erin, Tony \& Ram- as well as external colleagues -Aleks Jakulin from Data Flowers- for their input.

\printbibliography
\section*{Appendix}
\textbf{Passage 1}:

\textit{Computational Complexity Theory is a fundamental field within theoretical computer science that focuses on classifying computational problems according to their inherent difficulty and determining the resources needed to solve them. This theory provides a framework for understanding the limits of what can be achieved with algorithms and computation. The importance of this field stems from its implications for all computational processes, from simple algorithms that run on personal computers to complex calculations on supercomputers.}

\textit{The origins of Computational Complexity Theory date back to the 1960s, when researchers began to formally investigate the efficiency of algorithms. The primary goal was to categorize problems based on the amount of computational resources, such as time and memory, required to solve them. This categorization led to the creation of complexity classes, such as P (Polynomial time), NP (Nondeterministic Polynomial time), and PSPACE (Polynomial Space), among others. Each class represents a set of problems based on the resources needed for their solution under specific computational models.}

\textit{One of the central questions in Computational Complexity Theory is the P vs NP problem, which asks whether every problem whose solution can be verified quickly (in polynomial time) can also be solved quickly. This question remains one of the most profound unsolved problems in computer science and has significant implications for mathematics, cryptography, and the philosophy of science. A solution to this problem would fundamentally alter our understanding of problem-solving capabilities in the computational realm.}

\textit{Another significant aspect of complexity theory is the study of algorithmic efficiency. This involves analyzing algorithms to determine the least amount of resources necessary to solve a problem and finding ways to optimize these algorithms. Techniques like approximation algorithms become crucial when dealing with NP-complete problems, where finding the optimal solution efficiently is unfeasible with current technology and known methods.}

\textit{Moreover, Computational Complexity Theory extends beyond theoretical interest. Its practical applications include optimizing tasks in computer networks, improving software performance, and securing data through cryptographic protocols, which rely on the complexity of certain problems to ensure security.}

\textit{In conclusion, Computational Complexity Theory not only enriches our understanding of computation but also challenges and enhances our problem-solving capabilities across various scientific and technological domains. As computational devices become increasingly integral to solving complex global challenges, the insights from this theory will be crucial in paving the way for more efficient and effective solutions. This field continues to be vibrant and dynamically evolving, consistently pushing the boundaries of what can be achieved with computation.}

\textbf{Passage 2}:

\textit{General Relativity, a profound scientific theory developed by Albert Einstein in 1915, stands as one of the pillars of modern physics. It revolutionized our understanding of gravity, replacing Isaac Newton's theory that had reigned supreme for over two centuries. General Relativity is not merely a theory about gravity, however; it is a comprehensive framework for describing the fundamental structure of the universe.}

\textit{At the heart of General Relativity is the startling idea that gravity is not a conventional force acting in spacetime, but rather a manifestation of spacetime itself being curved by mass and energy. This notion diverges radically from Newtonian physics, where gravity was seen as an attractive force between masses. Einstein proposed that massive objects warp the fabric of spacetime, and this curvature dictates the motion of objects. This means that the Earth orbits the Sun not because it is being "pulled" by the Sun in a direct manner, but because it is moving along a curved path defined by the Sun's warping of spacetime.}

\textit{One of the most elegant aspects of General Relativity is how it extends the principle of relativity to include all constant motion. Whether an object is moving at a steady velocity or standing still, the laws of physics are the same. Einstein’s equations — field equations — describe how energy and momentum dictate spacetime curvature. These equations are highly complex and typically require sophisticated mathematical tools like tensors and differential geometry for their expression and solutions.}

\textit{General Relativity not only expanded our understanding of the cosmos on a large scale, including planets, stars, and galaxies, but also predicted phenomena that were not observed until decades after Einstein’s initial publication. One of the most famous predictions is the bending of light by gravity, which was confirmed during the solar eclipse of 1919 when starlight passing near the Sun was observed to bend along the curvature of spacetime, just as General Relativity predicted.}

\textit{Further implications of General Relativity include the prediction of black holes, regions of space where the spacetime curvature becomes so extreme that nothing, not even light, can escape from it. Another intriguing prediction is gravitational waves, ripples in the fabric of spacetime that propagate as waves at the speed of light, generated by some of the most violent and energetic processes in the universe. These were directly detected a century later by the LIGO observatory in 2015, confirming yet another of Einstein’s predictions.}

\textit{General Relativity remains a central element of modern physics and astronomy, essential not only for the theoretical frameworks it provides but also for practical applications. For instance, the theory is crucial for the calculations that enable the GPS technology found in smartphones and cars; without corrections based on General Relativity, GPS would fail in its navigational accuracies within minutes.}

\textit{Thus, General Relativity is not just a theory about how gravity works; it's a comprehensive description of the governing dynamics of the cosmos, influencing not only the trajectory of celestial bodies but also the everyday technologies that modern society depends on. Its beauty lies in its ability to explain the universe in terms of elegant mathematical models, making it one of the most beautiful and significant achievements in the history of human thought.}

\textbf{Passage 3}:
\textit{Neuroscience is a multidisciplinary science that is primarily concerned with the study of the structure and function of the nervous system and brain. The field spans several levels, from molecular to cellular to systems and cognitive neuroscience, providing insights into the complex mechanisms that underpin behavior and cognitive functions. As a science, neuroscience intersects with numerous fields such as psychology, computer science, statistics, pharmacology, and medicine, illustrating its broad relevance and application.}

\textit{The origins of neuroscience can be traced back to ancient Egypt, but significant strides have been made since the 20th century when it emerged as a distinct discipline. Today, neuroscience is guided by advanced imaging technologies like MRI (Magnetic Resonance Imaging) and PET (Positron Emission Tomography), which allow researchers to observe the brain's structure and functionalities in real-time and with high precision. These tools have revolutionized our understanding of various brain functions, including learning, memory, emotions, and sleep, along with the pathological bases of neurological disorders such as Parkinson's disease, Alzheimer's disease, and multiple sclerosis.}

\textit{At the cellular level, neuroscientists study neurons—the brain's primary functional units—and their complex networks known as neural circuits. By understanding how neurons communicate through neurotransmitters and synaptic connections, scientists are unraveling the biological bases of behaviors and how these processes go awry in mental health disorders.}

\textit{Systems neuroscience looks at how different neural circuits work together to perform complex brain functions. This area involves studying how the brain integrates information from the sensory organs, processes it, and responds. This knowledge is crucial for developing artificial neural networks in artificial intelligence applications and improving sensory prosthetics, such as cochlear implants for the deaf or retinal chips for the blind.}

\textit{Cognitive neuroscience bridges the gap between the study of the brain's physical structure and psychology by examining how cognitive processes such as thinking, memory, and attention are governed by neural circuits. This has profound implications for education, allowing for the development of learning methods that align with the brain's natural processes, and for healthcare, by providing new approaches to manage cognitive disorders and rehabilitation strategies.}

\textit{Furthermore, neuroscience research holds the key to the future of medical treatments. It is at the forefront of developing neuropharmacology, which could lead to new drugs for treating mental health disorders. Gene therapy and brain-computer interfaces represent other exciting prospects, potentially offering remedies or enhanced abilities for those suffering from brain injuries or degenerative brain diseases.}

\textit{In conclusion, neuroscience is not just an academic discipline but a crucial field that has the potential to profoundly impact various aspects of human life and health. It combines theoretical research with practical applications, aiming to answer fundamental questions about human nature while addressing some of the most challenging medical conditions. Its future, rich with the promise of technological and therapeutic advancements, is as exciting as it is essential.}

\textbf{Passage 4}:

\textit{String theory, an ambitious theoretical framework, aims to unify the seemingly incompatible theories of quantum mechanics and general relativity into a single, coherent model of the fundamental forces and particles in the universe. Since its emergence in the late 20th century, string theory has captivated physicists and mathematicians alike with its elegant approach and profound implications, despite its complex mathematical underpinnings and lack of direct experimental evidence.}

\textit{At the heart of string theory is the radical idea that the point-like particles which make up the fabric of our universe, such as electrons and quarks, are not point-like at all. Instead, these fundamental particles are conceived as tiny, vibrating strings. The frequency and mode of these vibrations determine the type of particle they manifest as, much like different musical notes arise from different vibrations of guitar strings. This simple yet profound shift from point particles to strings helps to avoid some of the problematic infinities that arise in quantum field theories when trying to describe the gravitational forces.}

\textit{String theory is not just about strings. Over time, it has evolved to include higher-dimensional objects known as branes, which are generalizations of strings that can have multiple dimensions. These branes interact in 10 or more dimensions, far beyond the four-dimensional spacetime framework established by Einstein's theory of relativity. The extra dimensions proposed by string theory are typically compactified or curled up in such a way that they are imperceptible at human scales, making experimental verification challenging.}

\textit{One of the most tantalizing aspects of string theory is its potential to be a "theory of everything." It aims to provide a single framework that encompasses all fundamental forces, including gravity, electromagnetism, the weak nuclear force, and the strong nuclear force. In doing so, string theory hopes to explain how these forces interact at the most fundamental level, potentially leading to insights about the earliest moments of the universe, including the Big Bang and the unification of forces at high energies.}

\textit{ Despite its beauty and promise, string theory faces significant challenges. The lack of direct empirical evidence and the extremely high energy scales involved, which are currently beyond the reach of our most powerful particle accelerators, make it difficult to test the predictions of string theory. Additionally, the mathematical complexity of the theory itself means that only a handful of specialists truly understand the intricacies and nuances of the full theory.}

\textit{In conclusion, string theory remains one of the most fascinating and challenging areas of modern theoretical physics. It offers a rich tapestry of ideas that intertwines the concepts of different dimensions, fundamental particles, and forces into a single narrative. Whether string theory will ultimately succeed as a theory of everything is still an open question, but its contribution to pushing the boundaries of human knowledge and its role in stimulating discussions across various disciplines of physics cannot be underestimated. As research continues, it may either revolutionize our understanding of the universe or serve as a stepping stone to even more comprehensive theories.}

\textbf{Passage 5}:

\textit{As you would have received in the email, currently, ACH transters, wire transfers, and card transactions are experiencing interruptions due to our bank partner Evolve Bank \& Trust. Unfortunately this happened without any prior notification to us. It seems other partners of Evolve Bank \& Trust are facing a similar issue as well.While we work to resume these services with our banking services provider, we understand that this must be frustrating for you. We would like to emphasize that Juno is not responsible for these disruptions and it's the responsibility of the banking services provider to provide these services and ensure its continuity. Generally, Consumer Financial Protection Bureau takes such banking disruptions without prior notice very seriously and takes necessary steps to ensure that the banking services provider remediates the issue swiftly.}

\textit{We would like to thank you for your patience and understanding as we fix the issues.}

\end{document}